\documentclass{article}

\usepackage[nonatbib,preprint]{nips_2018}
\usepackage{rotating}
\newcommand*\rot{\rotatebox{90}}
\usepackage{booktabs}
\usepackage{amsmath}
\usepackage{amssymb}
\usepackage{graphicx}
\usepackage{subcaption}
\usepackage{algorithm}
\usepackage{algorithmic}
\usepackage{enumitem}
\usepackage{multirow}
\usepackage{wrapfig}
\setlist{leftmargin=2.5mm}

\title{Domain Stylization:
A Strong, Simple Baseline for Synthetic to Real Image Domain Adaptation}

\author{
	Aysegul Dundar, Ming-Yu Liu, Ting-Chun Wang, John Zedlewski, Jan Kautz\\
	NVIDIA\\
	\texttt{\{adundar,mingyul,tingchunw,jzedlewski,jkautz\}@nvidia.com} \\
}

\begin{document}

\maketitle

% You may provide any keywords that you
% find helpful for describing your paper; these are used to populate
% the "keywords" metadata in the PDF but will not be shown in the document

% this must go after the closing bracket ] following \twocolumn[ ...

% This command actually creates the footnote in the first column
% listing the affiliations and the copyright notice.
% The command takes one argument, which is text to display at the start of the footnote.
% The \icmlEqualContribution command is standard text for equal contribution.
% Remove it (just {}) if you do not need this facility.

%\printAffiliationsAndNotice{}  % leave blank if no need to mention equal contribution
%fid_analysis\printAffiliationsAndNotice{} % otherwise use the standard text.

\begin{abstract}

Deep neural networks have largely failed to effectively utilize synthetic data when applied to real images due to the covariate shift problem. In this paper, we show that by applying a straightforward modification to an existing photorealistic style transfer algorithm, we achieve state-of-the-art synthetic-to-real domain adaptation results. We conduct extensive experimental validations on four synthetic-to-real tasks for semantic segmentation and object detection, and show that our approach exceeds the performance of any current state-of-the-art GAN-based image translation approach as measured by segmentation and object detection metrics. Furthermore we offer a distance based analysis of our method which shows a dramatic reduction in Frechet Inception distance between the source and target domains, offering a quantitative metric that demonstrates the effectiveness of our algorithm in bridging the synthetic-to-real gap.

%Furthermore we offer a Frechet Inception distance based analysis of our algorithm which shows the resulting translated images by our approach are much closer 

%domain adaptation improvement

%each step of our proposed approach by showing improved quantitative metrics in both stages of our algorithm.

% Performance of deep neural networks trained with synthetic images degrades when they are applied to real images due to the covariate shift problem. In this paper, we show that, by applying a simple tweak to an off-the-shelf photorealistic style transfer algorithm, one can effectively reduce the degradation. We conduct extensive experimental validation on $4$ synthetic-to-real tasks for semantic segmentation and object detection. The results show that our approach exceeds the performance of current state-of-the-art GAN-based image translation approaches.
% Deep neural networks have largely failed to effectively use synthetic data when applied to real images due to the covariate shift problem. In this paper, we show that by applying a straightforward modification to an existing photorealistic style transfer algorithm, we achieve state of the art synthetic-to-real domain adaptation results. We conduct extensive experimental validation on four synthetic-to-real tasks for semantic segmentation and object detection, and show that our approach exceeds the performance of any current state of the art GAN-based image translation approach.
\end{abstract}

\section{Introduction}

Training deep neural networks with Computer Graphics (CG) rendered images has recently emerged as a promising approach for various visual object recognition tasks. With CG engines, one can easily generate a huge number of labeled training images. While CG-rendered images seem to be the perfect data source, a deep neural network trained with synthetic images is often much less accurate than one trained with a much smaller number of real images due to the covariate shift problem~\cite{shimodaira2000improving}.

To address the covariate shift problem, several works propose learning a synthetic-to-real image translation function~\cite{taigman2016unsupervised,shrivastava2016learning,zhu2017unpaired,liu2017unsupervised,hoffman2017cycada}. These works are all based on the generative adversarial networks (GANs) framework~\cite{goodfellow2014generative}. Typically, they set up a zero sum game to be played by an image translation network and a discriminator, where the goal of the image translation network is to make a synthetic image look real so that the discriminator cannot tell it apart from real images. To ensure that a translated synthetic image shares the same semantic content, various constraints (e.g., cycle-consistency, shared latent space, or semantic consistency) are enforced. Using the translation function, one can use translated synthetic images to better train a deep neural network. 

We propose a novel method for training deep neural networks with synthetic images called domain stylization. It is based on an existing photorealistic image style transfer algorithm~\cite{li2018closed}. Our main contribution is to show that with a straightforward modification, we can make this stylization algorithm effective for the synthetic-to-real task. Specifically, we use this stylization algorithm to stylize synthetic images using randomly paired real images and propose using the resulting images to train deep neural networks (Section \ref{sec:training}). Experimental results on four synthetic-to-real tasks for semantic segmentation and object detection show that the proposed method achieves better accuracy in most of the scenarios, outperforming state-of-the-art image translation methods by a large amount (Section \ref{sec::exp}). Our second contribution is to provide a Frechet Inception Distance based analysis illustrating how our method mitigates the covariate shift problem (Section \ref{sec::disc}). 

Our approach has several advantages over previous approaches. First, we do not need to train an image-to-image translation network. While GANs are notoriously hard to train, our approach does not require any training. Second, the stylization process can potentially be done ``on the fly,'' which means we do not need to pre-stylize all the synthetic images, thus saving much effort and space. Finally, our method achieves better accuracy in most of the scenarios, outperforming state-of-the-art image translation methods by a large margin.

\section{Related Works}

\textbf{Learning with synthetic images.} Training machine learning models using CG-rendered depth images is a well-established practice for depth-based object recognition~\cite{shotton2011real,sharp2015accurate,peng2015learning,richter2015discriminative}. However, the same success has not yet been extended to CG-rendered RGB images due to challenges in rendering realistic RGB images. 

\textbf{Synthetic-to-real image translation.} Recently, several works utilize GANs~\cite{goodfellow2014generative} for image-to-image translation~\cite{isola2016image,wang2017high,taigman2016unsupervised,shrivastava2016learning,zhu2017unpaired,liu2017unsupervised,huang2018multimodal}. By translating an image from synthetic to real domains, one can make CG-rendered images more realistic. Based on the observation that existing image-to-image translation methods have no guarantee that a translated synthetic image preserves the original semantic layout, Hoffman et.~al.~\cite{hoffman2017cycada} incorporate a semantic layout preserving loss in image-to-image translation. These works show promising results on the synthetic-to-real task but have to deal with the difficult problem of GAN training. In our work, we take a different avenue to make synthetic images more realistic. Our approach is based on randomly pairing a synthetic image to a real image and transferring the style in the real image to the synthetic image. The style transfer is achieved by simple, efficient feature transform and image smoothing operations. Our experimental results show that the stylized synthetic images yield a better synthetic-to-real performance.

\textbf{Domain adaptation.} Synthetic-to-real is a special case of the visual domain adaptation problem~\cite{saenko2010adapting}, which concerns adapting a classifier trained in one visual domain to another. Various methods exist, ranging from metric learning~\cite{kulis2011you,gong2012geodesic}, subspace modeling~\cite{gopalan2011domain,fernando2013unsupervised}, deep feature alignment~\cite{tzeng2014deep}, to adversarial training~\cite{ganin2016domain,liu2016coupled}.

\section{Domain Stylization} \label{sec:training}

We propose Domain Stylization (\texttt{DS}), an effective approach of using the existing FastPhotoStyle~\cite{li2018closed} to generate stylized synthetic image datasets for the synthetic-to-real problem. We generate the stylized synthetic dataset as follows:
Let $D^{S}=\{(\mathbf{x}_i^{S},\mathbf{m}_i^{S})\}$ be a synthetic image dataset that consists of pairs of CG-rendered image $\mathbf{x}_i^{S}$ and its rendered semantic segmentation mask $\mathbf{m}_i^{S}$. Let $D^{R}=\{(\mathbf{x}_j^{R},\tilde{\mathbf{m}}_j^{R})\}$ be a real image dataset that consists of pairs of real image $\mathbf{x}_i^{R}$ and its \emph{estimated} semantic segmentation mask $\tilde{\mathbf{m}}_i^{R}$. (We will discuss how to obtain these estimated semantic segmentation masks below.) To create the stylized synthetic dataset, for every image--label pair in $D^{S}$, we randomly sample $N$ image--label pairs in $D^{R}$ to generate $N$ stylized synthetic images. For example, when picking the $i$th and $j$th pairs from $D^S$ and $D^R$, respectively, the output is given by $(\mathbf{x}_{ij}^{S},\mathbf{m}_i^{S})$ where $\mathbf{x}_{ij}^{S}=f(\mathbf{x}_i^{S},\mathbf{m}_i^{S},\mathbf{x}_j^{R},\tilde{\mathbf{m}}_j^{R})$. Note that the segmentation mask remains the same since the FastPhotoStyle method does not change semantic content in an image. %The pseudo-code of this procedure is given in the supplementary material.

Visual quality of the stylization output depends on the selection of the style image. A randomly selected style image could lead to an undesired output. Ideally, one needs to search for the best style image in the real dataset for each synthetic image. However, this is computationally expensive. Moreover, evaluating the visual quality of a stylization often requires human-in-the-loop, which is difficult to scale. Therefore, we use $N$ randomly selected style images in our implementation. Although this seems to be ad hoc, it works really well in practice. In our experiments, we set $N=10$ via an ablation study. We also note that there exist cases where a semantic label in the synthetic image is missing in the real image. In this case, the FastPhotoStyle method will only stylize those regions where a matched region exists and leave the style in the unmatched region unchanged. These cases are non-ideal but they do not cause problems in practice.

\begin{figure}[tbh!]
\centering
\includegraphics[width=1.0\textwidth]{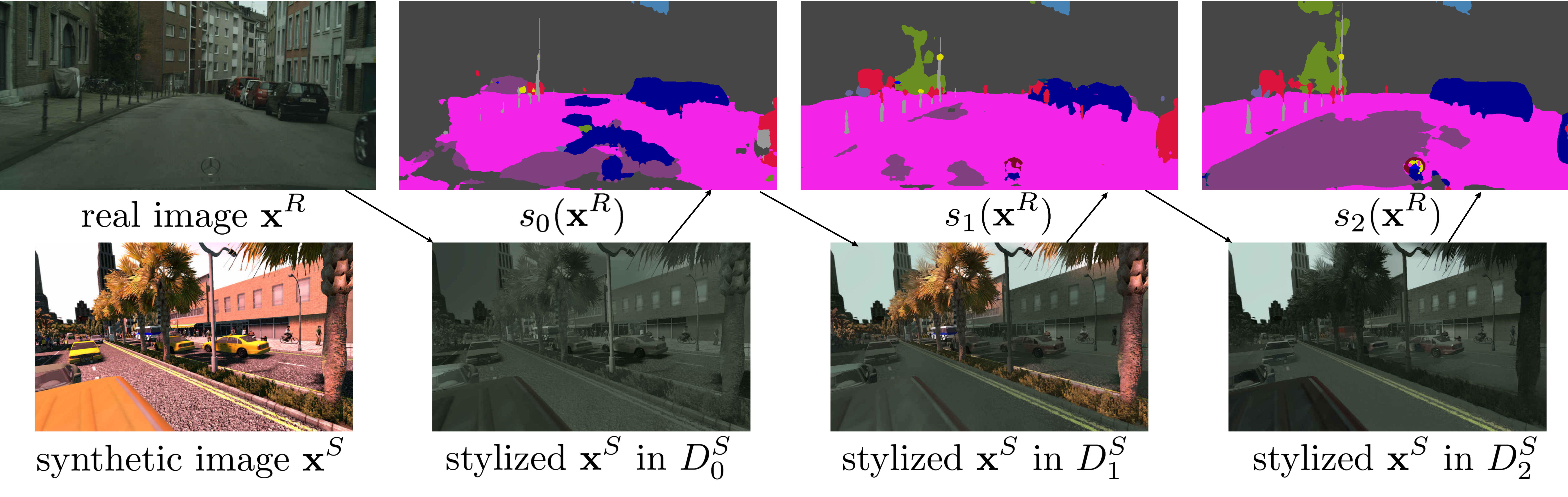}
\vspace{-3mm}
\caption{Examples of intermediate stylization outputs. First, the synthetic image is stylized using the real image ($x^R$) without the aid of any segmentation map. Then we train a segmentation network ($s_0$) using these stylized images, and use $s_0$ to segment the real images ($s_0(x^R)$). Given the segmentation map, we can then better stylize the synthetic image and so on.}
\label{fig::algorithm_flow}
%\vspace{-3mm}
\end{figure}

\begin{algorithm}[tb!]
  \caption{Training with stylized synthetic datasets.}
  \label{alg:training-with-domain-stylization}
\begin{algorithmic}
\begin{small}
  \STATE {\bfseries Input:} $D^{S}=\{(\mathbf{x}_i^{S},\mathbf{m}_i^{S})\}$, $D^{R}=\{(\mathbf{x}_j^{R})\}$, $T$, $N$
  \STATE Compute $D_0^S=\text{\texttt{DS}}    (\{(\mathbf{x}_i^{S},\mathbf{1})\},\{(\mathbf{x}_j^{R},\mathbf{1})\}, N)$
  \FOR{$t=0$ {\bfseries to} $T$}
  \STATE Compute $s_t=\text{\texttt{SSL}}(D_t^S)$
  \STATE Generate $D_{t+1}^S=\text{\texttt{DS}}    (D^{S},\{(\mathbf{x}_j^{R},s_t(\mathbf{x}_j^{R})\},N)$
  \ENDFOR
  \STATE {\bfseries Output:} $S_T$
\end{small}
\end{algorithmic}
\end{algorithm}

We now describe how we use the stylized synthetic dataset for the semantic segmentation task. Our method is an iterative approach. It iterates between two steps: the Domain Stylization (\texttt{DS}) step and the Semantic Segmentation Learning (\texttt{SSL}) step (referring to the training procedure for training a semantic segmentation network). Since semantic segmentation masks for synthetic images can be easily computed by using CG engines (e.g., via embedding object class ID in the material map), we hence assume they are always available. We use the semantic segmentation network learned in the intermediate step for computing the semantic segmentation masks for the real images.

We first generate a stylized synthetic dataset without using segmentation masks. This is achieved by assigning all pixels to 1 in the segmentation masks:
\begin{equation}
    D_0^S=\text{\texttt{DS}}    (\{(\mathbf{x}_i^{S},\mathbf{1})\},\{(\mathbf{x}_j^{R},\mathbf{1})\})
\end{equation}
We then use $D_0^S$ to train a semantic segmentation network, $s_0$. The trained semantic segmentation network is then used to compute the semantic segmentation mask for each real images, $s_0(\mathbf{x}_j^{R})$. The synthetic dataset is then stylized using the real image dataset with the new segmentation masks for creating a new stylized synthetic dataset
\begin{equation}
    D_1^S=\text{\texttt{DS}}    (\{(\mathbf{x}_i^{S},\mathbf{m}_i^{S})\},\{(\mathbf{x}_j^{R},s_0(\mathbf{x}_j^{R}))\})
\end{equation}
The new stylized dataset is used to train a new segmentation network $s_1$. We iterate these two steps for a number of iterations, $T$. We set $T=2$ in our experiments via an ablation study. The learned segmentation network $s_T$ from the last iteration is the output, which we use to segment unseen real images. The procedure is summarized in Algorithm \ref{alg:training-with-domain-stylization}. In Figure~\ref{fig::algorithm_flow}, we show the intermediate segmentation masks and stylized synthetic images for a pair of synthetic and real images. When conducting benchmark evaluation, we use the defined training set (images only) for stylization and report performance on the test set.

%\vspace{-.1in}
\section{Experiments}\label{sec::exp}
%\vspace{-.05in}

In this section, we will first evaluate our method on synthetic-to-real street scene segmentation and synthetic-to-real indoor scene segmentation. We will then present an ablation study analyzing several design choices in the proposed method. Finally, we also apply the proposed algorithm to a synthetic-to-real object detection task. For all the experiments, we compare the proposed method to several recent synthetic-to-real methods, including GAN-based image translation methods and domain randomization. We follow the synthetic-to-real settings in the prior works where we treat the synthetic dataset as the source domain and the real dataset as the target domain. Ground truth semantic segmentation labels from the real datasets is neither used for stylization nor for training. %We will opensource our implementation.

%%%%%%%%%%%%%%%%%%%%%%%%%%%%%%%%%%%%%%%%%%%%%%%%%%%%%%%%%%%%%%%%%%%%%%%%

%\vspace{-.02in}
\subsection{Synthetic-to-real Street Scene Segmentation}
%\vspace{-.02in}

\begin{figure}[tbh!]
\centering
\includegraphics[trim={0 0 0 1cm},clip, width=0.99\textwidth]{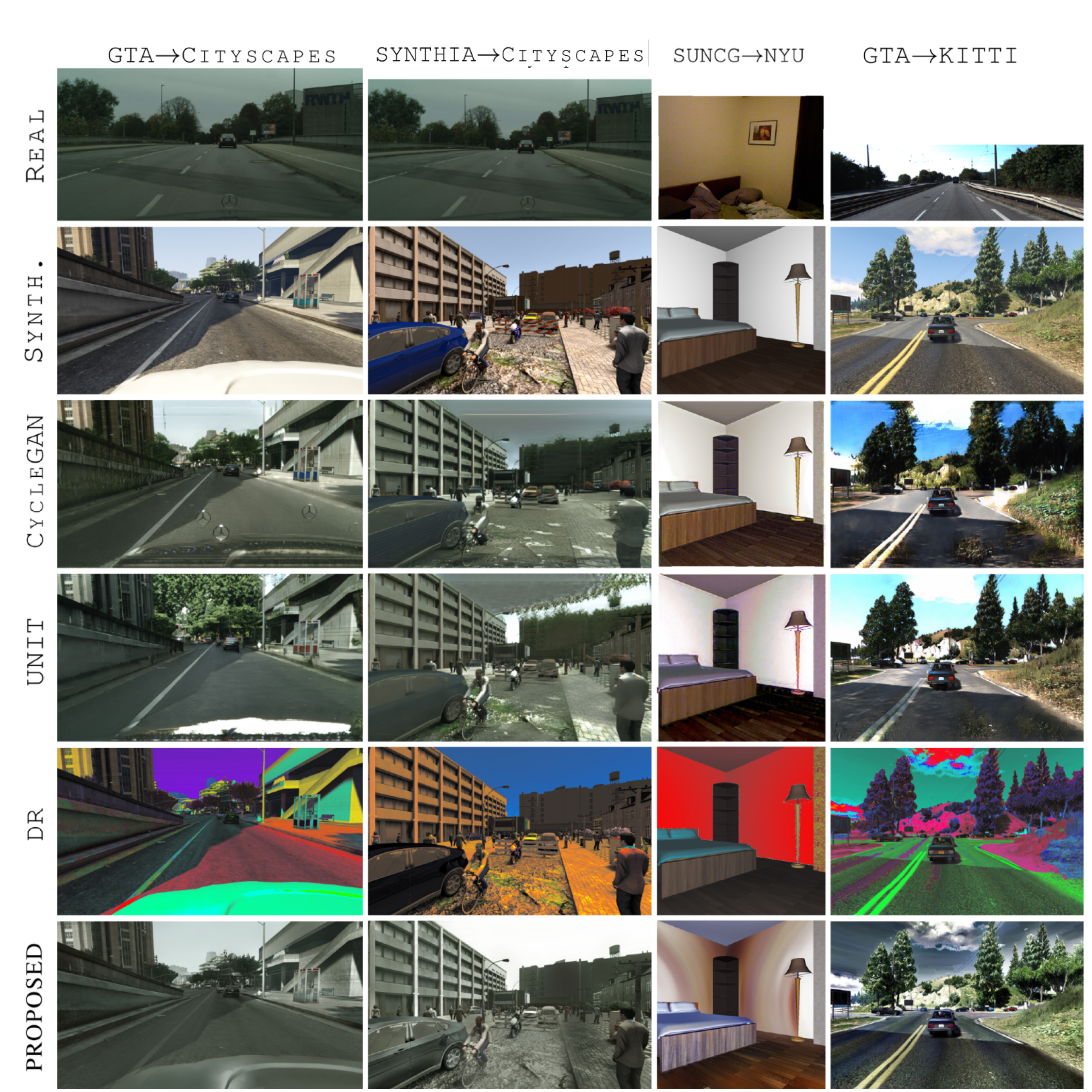}
%\vspace{-2mm}
\caption{Example images of target, source and generated images by competing methods and domain stylization. }
\label{fig::example}
%\vskip -0.2in
\end{figure}

{\bf Datasets.} For synthetic-to-real street scene segmentation, the evaluation is based on the Cityscapes benchmark~\cite{cordts2016cityscapes}. The benchmark consists of urban street scene images from various European cities and consider 19 semantic object classes, including road, sky, car, pedestrian, etc. We use the Cityscapes training set, which consists of 2,975 images %from 18 cities, 
for stylizing synthetic images.
We use two different synthetic datasets separately. The first one is the GTA dataset provided by~\cite{richter2016playing}. It consists of 24,966 images rendered from the Grand Theft Auto V video game. The second synthetic dataset is the SYNTHIA dataset~\cite{ros2016synthia}, which consists of 9,400 images rendered from driving a virtual car in several virtual cities.
%under different illumination and weather conditions.
Semantic segmentation ground truth for the two synthetic datasets are available. We train semantic segmentation networks using the stylized versions of these synthetic datasets and evaluate the performance using the Cityscapes validation set.
%, which consists of 500 images from 3 different cities. Note that the Cityscapes benchmark splits the training and validation sets based on cities, so there is no overlap between the cities in the training sets and those in the validation set.
In summary, we evaluate the following two synthetic-to-real settings: \texttt{GTA}$\longrightarrow$\texttt{Cityscapes} and \texttt{SYNTHIA}$\longrightarrow$\texttt{Cityscapes}.

{\bf Semantic segmentation network.} We use the $26$-layer Dilated Residual Network \texttt{DRN-C-26} \cite{yu2017dilated} for semantic segmentation due to its good performance. The network is based on ResNet~\cite{he2015deep} and dilated convolution~\cite{yu2015multi,chen2015semantic}. We use the source code provided by the authors for the experiments. Specifically, we finetune from the ImageNet-pretrained model using \texttt{SGD} with learning rate $0.001$ and momentum $0.99$. We decay the learning rate by a factor of $10$ for every $100$ epochs and train the network for $250$ epochs in total. For a fair comparison, we evaluate all the competing methods using \texttt{DRN-C-26} with the same training hyperparameters. We report performance achieved in the last epoch for all the competing methods.

{\bf Competing methods.} We compare the proposed method to various synthetic-to-real methods. We use the implementations provided by the authors to generate training data for training the networks. Some example images generated by the algorithms are shown in Figure~\ref{fig::example}.

\begin{itemize}[noitemsep]
% %\vspace{-8mm}
\item \texttt{Real.} The network is trained using the real image training set; the result is served as a performance upper-bound.
\item \texttt{Synth.} In this setting, we directly train the network using synthetic images in their original version. 
\item \texttt{Domain Randomization} \cite{tobin2017domain} aims at generalizing a model, trained purely using synthetic data, to recognize real world cases by providing sufficient simulated variability at training time. Its effectiveness was shown on a task of detecting objects of simple geometry (e.g., cylinders and cubes) for a robotic grasping task. In the original implementation, the variety is achieved by pasting random textures on object meshes where the random textures are created with a random RGB value, a gradient between two random RGB values, or a checkerboard pattern of two random RGB values. In order to deal with complex real-world street scenes, we implement the domain randomization idea by randomly changing the color of each object in a rendered scene. Specifically, we use the HSV color representation and apply a random shift to the hue value\footnote{The shift is a circular shift because he hue value is in $[0\medspace360]$.} of an object where the object region is identified using the segmentation mask. We will refer to this method as \texttt{DR}. %Examples of the domain randomized images are shown in Figure \ref{cityscapes-examples}(e). 

\item \texttt{CycleGAN} \cite{zhu2017unpaired} is an image-to-image translation model. It is based on GAN~\cite{goodfellow2014generative} and a cycle-consistency constraint. We train \texttt{CycleGAN} to translate images between the synthetic and real image datasets, and use the obtained network to translate a synthetic image to a corresponding real image. We then use the translated synthetic images to train the semantic segmentation network. We resize the images in both the real and synthetic datasets so that the longer side of the image is 1024 for easing the adversarial learning. The same processing is applied to the other image translation-based synthetic-to-real methods.

\item \texttt{UNIT} \cite{liu2017unsupervised} is also an image-to-image translation model. Unlike \texttt{CycleGAN}, it is based on the Coupled GAN framework~\cite{liu2016coupled}, which assumes a shared-latent space between two image domains. We apply \texttt{UNIT} to translate a synthetic image to a real image for training the semantic segmentation network.

\item \texttt{CyCADA} \cite{hoffman2017cycada} is an extension of the \texttt{CycleGAN} algorithm where a semantic layout preserving loss is incorporated to ensure the translation preserves the original image content.

\item \texttt{FCNs}-\texttt{ITW} \cite{hoffman2016fcns} is based on feature space adversarial training for learning domain invariant feature representations.
\end{itemize}
%%\vspace{-4mm}

\paragraph{Results.} In Table~\ref{fig::cityscapes-table}, we report the \texttt{GTA}$\longrightarrow$\texttt{Cityscapes} performance achieved by various competing algorithms where per-class IoU, mean IoU, and pixel accuracy numbers are shown. We find that the proposed algorithm achieved the best pixel accuracy score of 87.2. For mean IoU, \texttt{CycleGAN}'s performance is 39.6, which is slightly better than 38.3 achieved by the proposed algorithm. We note that the performance numbers of \texttt{CyCADA} and \texttt{FCNs}-\texttt{ITW} are duplicated from the original papers where the base network of \texttt{CyCADA} is identical to ours.

The results on \texttt{SYNTHIA}$\longrightarrow$\texttt{Cityscapes} is also reported in Table~\ref{fig::cityscapes-table}. We find that the proposed algorithm achieves a mean IoU score of 29.5 and a pixel accuracy score 76.5, which are significantly better than those achieved by the other competing algorithms.

\begin{table*}[t]
\caption{Results on Cityscapes validation set; IoU for each class, mean IoU, and pixel accuracy of different adaptation methods between GTA $\rightarrow$ Cityscapes and SYNTHIA $\rightarrow$ Cityscapes. }
%CyCADA \cite{hoffman2017cycada}
\label{fig::cityscapes-table}

%\begingroup
\centering
\renewcommand\tabcolsep{1.2pt}
\begin{scriptsize}
\begin{sc}
\begin{tabular}{l|ccccccccccccccccccc|c|c|}
\toprule
 \multicolumn{22}{c}{\texttt{GTA}$\rightarrow$\texttt{Cityscapes}} \\
\toprule
 & \rot{road} & \rot{sidewalk}	 & \rot{building} &	\rot{wall} & \rot{fence} &	\rot{pole} &	\rot{traffic light} &	\rot{traffic sign} &	\rot{vegetation} &	\rot{terrain} &	\rot{sky} &	\rot{person} &	\rot{rider} &	\rot{car} & \rot{truck} &
\rot{bus} &	\rot{train} &	\rot{motorbike} &	\rot{bicycle} &	\rot{mean IoU} &	\rot{Pixel Acc.} \\
\midrule
\texttt{Synth.} & 68.9& 19.9& 52.8& 6.5& 13.6& 9.3& 11.7& 8.0& 75.0& 11.0& 56.5& 36.9& 0.1& 51.3& 8.5& 4.7& 0.0& 0.1& 0.0& 22.9& 71.9 \\
\texttt{DR} & 67.5& 23.5& 65.7& 6.7& 12.0& 11.6& 16.1& 13.7& 70.3& 8.3& 71.3& 39.6& 1.6& 55.0& 15.1& 3.0& 0.6& 0.2& 3.3& 25.5& 73.8 \\
\texttt{CycleGAN} & 89.3& \textbf{45.1}& \textbf{81.6}& \textbf{27.5}& \textbf{18.6}& 29.0& \textbf{35.7}& 17.3& 79.3& \textbf{29.4}& 71.5& 59.7 & 15.7& \textbf{85.3}& 18.2& 14.8& 1.4& \textbf{21.9}& 12.5& \textbf{39.6}& 86.6 \\
\texttt{UNIT} & \textbf{90.5}& 38.5& 81.1& 23.5& 16.3& 30.2& 25.2& 18.5& 79.5& 26.8& 77.8& 59.2& \textbf{17.4}& 84.4& \textbf{22.2}& 16.1& 1.6& 16.7& 16.9& 39.1& 87.1  \\
\texttt{FCNs ITW} & 70.4& 32.4& 62.1& 14.9& 5.4& 10.9& 14.2& 2.7& 79.2& 21.3& 64.6& 44.1& 4.2& 70.4& 8.0& 7.3& 0.0& 3.5& 0.0& 27.1& - \\
\texttt{CyCADA} & 79.1& 33.1& 77.9& 23.4& 17.3& \textbf{32.1}& 33.3& \textbf{31.8}& 81.5& 26.7& 69.0& \textbf{62.8}& 14.7& 74.5& 20.9& 25.6& \textbf{6.9}& 18.8& \textbf{20.4}& 39.5& 82.3 \\
proposed & 89.0& 43.5& 81.5& 22.1& 8.5& 27.5& 30.7& 18.9& \textbf{84.8}& 28.3& \textbf{84.1}& 55.7& 5.4&  83.2& 20.3&	\textbf{28.3}& 0.1& 8.7& 6.2& 38.3& \textbf{87.2}  \\
\toprule
\multicolumn{22}{c}{\texttt{SYNTHIA}$\rightarrow$\texttt{Cityscapes}} \\
\toprule
\texttt{Synth.} & 28.5& 10.8& 49.6& 0.2& 0.0& 18.5& 0.7& 5.6& 65.3& 0.0& 71.6& 36.6& 6.4& 43.8& 0.0& 2.7& 0.0& 0.8& 10.0& 18.5& 54.6\\
\texttt{DR}& 31.3& 16.7& 59.5& 2.2& 0.0& 19.7& 0.4& 6.2& 64.7& 0.0& 67.3& 43.1& 3.9& 35.1& 0.0& 8.3& 0.0& 0.3& 5.5& 19.2& 57.9  \\
\texttt{CycleGAN}& 58.8& 20.4& 71.6& 1.6& \textbf{0.7}& 27.9& 2.7& 8.5& 73.5& 0.0& 73.1& 45.3& 16.2& 67.2& 0.0& 14.9& 0.0& 7.9& 24.7& 27.1& 71.4  \\
\texttt{UNIT}& 56.3& 20.6& 73.2& 1.8& 0.3& 29.0& 4.0& 11.8& 72.2& 0.0& 74.5& 50.7& \textbf{18.4}& 67.3& 0.0& \textbf{15.1}& 0.0& 6.7& \textbf{29.5}& 28.0& 70.8  \\
\texttt{FCNs ITW} & 11.5& 19.6& 30.8& \textbf{4.4}& 0.0& 20.3& 0.1& 11.7& 42.3& 0.0& 68.7& 51.2& 3.8& 54.0& 0.0& 3.2& 0.0& 0.2& 0.6& 17.0& -\\
proposed& \textbf{67.0}& \textbf{28.0}& \textbf{75.3}& 4.0& 0.2& \textbf{29.9}& \textbf{3.8}& \textbf{15.7}& \textbf{78.6}& 0.0& \textbf{78.0}& \textbf{54.0}& 15.4& \textbf{69.7}& 0.0& 12.0& 0.0& \textbf{9.9}& 19.2  & \textbf{29.5}& \textbf{76.5} \\
\midrule
\texttt{Real}&  97.2&	78.4&	90.0&	45.4&	45.4&	49.3&	52.7&	66.2&	90.2&	56.6&	92.9&	72.4&	48.9&	91.7&	52.7&	68.6&	51.3&	42.1&	67.6&	66.3&	94.3    \\
\bottomrule
\end{tabular}
%\endgroup
\end{sc}
\end{scriptsize}
\end{table*}

We repeat the experiments on these two datasets with a more powerful semantic segmentation network, \texttt{DeepLabV3} \cite{chen2017rethinking}. We see a similar accuracy improvement and consistent ranking between the algorithms. Results can be found in Table~\ref{fig::cityscapes-deeplab-table}.

\begin{table*}[tbh!]
	\caption{Results on Cityscapes validation set with a stronger semantic segmentation network.}
	\label{fig::cityscapes-deeplab-table}
	\centering
	\renewcommand\tabcolsep{1.2pt}
	\begin{scriptsize}
		\begin{sc}
			\begin{tabular}{l|ccccccccccccccccccc|c|c|}
				\toprule
				\multicolumn{22}{c}{\texttt{GTA}$\rightarrow$\texttt{Cityscapes} based on DeepLabV3} \\
				\toprule
				& \rot{road} & \rot{sidewalk}	 & \rot{building} &	\rot{wall} & \rot{fence} &	\rot{pole} &	\rot{traffic light} &	\rot{traffic sign} &	\rot{vegetation} &	\rot{terrain} &	\rot{sky} &	\rot{person} &	\rot{rider} &	\rot{car} & \rot{truck} &
				\rot{bus} &	\rot{train} &	\rot{motorbike} &	\rot{bicycle} &	\rot{mean IoU} &	\rot{Pixel Acc.} \\
				\midrule
				\texttt{Synth.} & 79.2& 26.9& 79.5& 19.1& 27.4& 29.4& 34.9& 17.9& 81.8& 25.3& 71.9& 60.8& 22.1& 81.5& 34.1& 17.0& 2.5& 21.3& 26.9& 40.0& 82.9  \\
				\texttt{CycleGAN} &  \textbf{89.9}& \textbf{45.4}& \textbf{85.6}& 38.3& 32.9& \textbf{36.1}& \textbf{45.1}& 27.6& 83.7& 28.9& 83.4& 65.8& \textbf{32.0}& \textbf{86.2}& 30.0& \textbf{32.1}& 1.8& \textbf{33.5}& 37.2& \textbf{48.2}& \textbf{88.4}\\
				\texttt{UNIT} &  87.6& 39.4& 85.2& 38.2& \textbf{35.1}& 35.3& 42.3& \textbf{33.4}& 76.0& 21.0& 69.4& 63.7& 29.3& 85.4& 34.1& 31.4& \textbf{28.5}& 31.5& \textbf{39.8}& 47.7& 86.0 \\
				proposed & 86.9& 44.5& 84.7& \textbf{38.8}& 26.6& 32.1& 42.3& 22.5& \textbf{84.7}& \textbf{30.9}& \textbf{85.9}& \textbf{67.0}& 28.1& 85.7& \textbf{38.3}& 31.8& 21.5& 31.3& 24.6& 47.8& 87.6  \\
				\toprule
				\multicolumn{22}{c}{\texttt{SYNTHIA}$\rightarrow$\texttt{Cityscapes} based on DeepLabV3}  \\
				\toprule
				\texttt{Synth.} & 66.0& 26.7& 77.4& 2.6& 0.2& 33.8& 16.0& 21.3& 77.9& 0.0& 81.8& 59.8& 23.0& 70.0& 0.0& 8.0& 0.0& 13.4& 26.3& 31.8& 76.7\\
				\texttt{CycleGAN}& 82.5& 38.5& 76.0& 0.7& 0.4& 32.2& 10.7& 18.4& 77.4& 0.0& 78.1& 60.2& \textbf{26.5}& \textbf{79.1}& 0.0& 23.0& 0.0& 12.1& 31.5& 34.0& 82.8 \\
				\texttt{UNIT}&  \textbf{86.2}& \textbf{43.9}& 76.6& 1.2& \textbf{1.4}& 33.4& 13.1& 20.4& 75.6& 0.0& 79.0& 57.8& 25.7& 77.3& 0.0& 23.0& 0.0& 17.0& \textbf{36.2}& 35.1& \textbf{84.2} \\
				
				proposed& 75.8& 32.2& \textbf{81.8}& \textbf{2.5}& 0.8& \textbf{35.9}& \textbf{20.6}& \textbf{23.3}& \textbf{81.5}& 0.0& \textbf{83.4}& \textbf{62.4}& 24.5& 78.0& 0.0& \textbf{28.2}& 0.0& \textbf{22.2}& 31.4& \textbf{36.0}& 81.8 \\
				\midrule
				\texttt{Real}&  97.9& 83.5& 91.6& 56.5& 61.2& 54.8& 63.9& 73.6& 91.3& 59.9& 93.2& 77.7& 60.1& 94.0& 79.3& 87.0& 76.1& 61.0& 73.2& 75.6& 95.5    \\
				\bottomrule
			\end{tabular}
			%\endgroup
		\end{sc}
	\end{scriptsize}
	%\end{center}
	%\vskip -0.1in
\end{table*}

\begin{table*}[tbh!]
\caption{Results on images that are generated with different image translation methods.}
\label{fig::cityscapes-training-table}

%\begingroup
\centering
\renewcommand\tabcolsep{1.2pt}
\begin{scriptsize}
\begin{sc}
\begin{tabular}{l|ccccccccccccccccccc|c|c|}
\toprule
 \multicolumn{22}{c}{\texttt{Cityscapes}$\rightarrow$\texttt{SYNTHIA}} \\
\toprule
 & \rot{road} & \rot{sidewalk}	 & \rot{building} &	\rot{wall} & \rot{fence} &	\rot{pole} &	\rot{traffic light} &	\rot{traffic sign} &	\rot{vegetation} &	\rot{terrain} &	\rot{sky} &	\rot{person} &	\rot{rider} &	\rot{car} & \rot{truck} &
\rot{bus} &	\rot{train} &	\rot{motorbike} &	\rot{bicycle} &	\rot{mean IoU} &	\rot{Pixel Acc.} \\
\midrule
\texttt{Synth.}& 39.8& 14.0& 64.4& \textbf{2.1}& 0.4& 12.7& \textbf{8.3}& 3.5& 47.8& 0.0& 76.2& \textbf{42.0}& 8.2& \textbf{29.0}& 0.0& \textbf{14.2}& 0.0& \textbf{10.1}& 11.0& 20.2& 58.9 \\
\texttt{CycleGAN} & 40.8& 13.2& 62.0& 0.5& 0.4& 11.3& 3.3& 3.9& 47.7& 0.0& 67.6& 17.0& 3.3& 16.3& 0.0& 2.3& 0.0& 4.2& 6.1& 15.8& 56.8 \\
\texttt{UNIT} &  43.7& 23.1& 58.2& 1.0& \textbf{0.9}& 13.8& 4.2& 5.8& 41.5& 0.0& 24.4& 29.0& 8.9& 20.0& 0.0& 3.7& 0.0& 5.8& 10.6& 15.5& 55.7 \\
proposed &  \textbf{46.6}& \textbf{23.8}& \textbf{65.8}& 1.5& 0.4& \textbf{17.2}& 6.5&  \textbf{6.2}& \textbf{53.5}& 0.0& \textbf{79.6}& 39.9& \textbf{11.2}& 25.9& 0.0& 7.7& 0.0& 7.1& \textbf{11.2}& \textbf{21.3}& \textbf{63.1} \\
\bottomrule
\end{tabular}
%\endgroup
\end{sc}
\end{scriptsize}
%\end{center}
%\vskip -0.1in
\end{table*}

In Table~\ref{fig::cityscapes-training-table}, we evaluate the network trained with Cityscapes images on the original SYNTHIA images and translated SYNTHIA images by competing methods and the proposed method. We again use the \texttt{DRN-C-26} architecture for this experiment.
We note that the accuracy degrades when the images are translated by UNIT and CycleGAN algorithms as these algorithms do no guarantee to preserve the semantic content. It can also be seen in Figure \ref{fig::example} that UNIT and CycleGAN sometimes convert sky to vegetation, remove parts of a bicycle etc., whereas the images that are translated with the proposed algorithm improve the results, and preserve the semantic content.

\subsection{Synthetic-to-real Indoor Scene Segmentation}

{\bf Datasets.} For synthetic-to-real indoor scene segmentation, the evaluation is based on the NYU-Depth V2 dataset~\cite{Silberman:ECCV12}. The NYU dataset comprises video sequences of various indoor scenes, recorded using both RGB and depth cameras. Full image semantic segmentation masks are available for a small subset of the images. We use 383 bedroom RGB images in the dataset for evaluation which are divided into a training set of 192 images and a validation set of 191 images. The image resolution is $640\times480$. We consider 6 semantic object classes including floor, wall, ceiling, bed, table, and chair. Pixels belong to the other semantic classes are ignored. We use the SUNCG dataset~\cite{song2016semantic}, which is a dataset of 45K indoor 3D scenes, to generate a dataset of synthetic bedroom images using the House3D~\cite{wu2018building} software. We generate 7,274 bedroom images from various 3D scenes. The image resolution is $640\times640$. The synthetic-to-real setting for the experiment is termed as \texttt{SUNCG}$\rightarrow$\texttt{NYU}.

\begin{table}[t]
\caption{Results on indoor scene segmentation.}
\label{tbl::nyu}
%%\vspace{-3mm}
\centering
\begin{scriptsize}

\begin{sc}
\begin{tabular}{l|c|c|c|c|c|c||c|c|}
\toprule
 \multicolumn{9}{c}{\texttt{SUNCG}$\rightarrow$\texttt{NYU}} \\
\toprule
 & \rot{floor} & \rot{ceiling}	 & \rot{wall} &	\rot{bed} & \rot{chair}  & \rot{table} &	\rot{meanIoU} &	\rot{Pxl Acc.} \\
\midrule
\texttt{Synth.}&   24.4& 4.0& 49.2& 20.3& 8.1& 1.5& 17.9& 49.6\\
\texttt{DR} & 9.0& 2.0& 29.4& 23.4& 5.2& 0.8& 11.6& 37.3 \\
\texttt{CycleGAN}& 23.7& \textbf{12.3}& 54.7& 27.4& 5.9& \textbf{4.0}& 21.3& 52.4 \\
\texttt{UNIT}& 14.0& 8.1& 62.7& 39.2 & 5.8& 0.5& 21.7& 60.0\\
proposed & \textbf{42.7}& 8.9& \textbf{76.9}& \textbf{51.3} & \textbf{20.2}& 1.9& \textbf{33.6}& \textbf{72.8} \\
\midrule
\texttt{Real}& 79.2& 19.2& 92.3& 76.8& 29.2& 5.9& 50.4& 89.3 \\
\bottomrule
\end{tabular}
%\endgroup
\end{sc}

\end{scriptsize}
\end{table}

{\bf Results.} As reported in Table~\ref{tbl::nyu}, we find that the proposed method achieves a mIoU of 33.6, which is 1.58 times better than 21.3 achieved by \texttt{CycleGAN}. In terms of pixel accuracy, the score of the proposed method is 72.8, outperforming 60.0 achieved by \texttt{UNIT}. 

\subsection{Ablation Study}

We analyze various algorithm design choices on the setting of  \texttt{SYNTHIA}$\longrightarrow$\texttt{Cityscapes}. We run each set-up with 5 random seeds and report the averaged results.

\textbf{Number of stylization $N$}. In Table \ref{tbl::ablation}, we report the performance of the proposed algorithm using different numbers of stylizations per synthetic image $N$. We find as increasing $N$ from 1 to 10 the performance scores increase a lot. But as we further increase $N$ to 25 or 50,  the performance scores saturate.

\textbf{Granularity of segmentation masks.} We analyze the impact of granularity of segmentation masks. We have $3$ different set-ups. The first one is not using any semantic segmentation mask; an example would be the stylized synthetic image at the $0$th iteration in Figure \ref{fig::algorithm_flow}. The second set-up is using a coarse semantic label mask which divides the regions into road, sky and others.  
Third set-up is using fine semantic label masks (19 classes), which corresponds to stylized synthetic image at the $1$st iteration in Figure~\ref{fig::algorithm_flow}. As shown in Table \ref{tbl::ablation}, stylizing synthetic image based on finer label masks leads to improved performance scores.

\textbf{Number of iterations $T$.} We iterate over stylization, training a network to estimate label masks, and stylization with the estimated label masks. We see an improvement of mean AP beginning with 27.1 ($T=0$), and increasing to 29.2 after the first iteration ($T=1$). With the second iteration ($T=2$), mAP increases to 29.5, and converges. The third iteration does not increase the accuracy further.

\begin{table}
\parbox{.4\linewidth}{

\caption{Ablation Study} \label{tbl::ablation}
%%\vspace{-4mm}
\begin{center}
\begin{scriptsize}
\begin{sc}
\begin{tabular}{l|c|c|c|}
\toprule
 &  Set-up & mean IOU & PXL ACC.\\

\midrule
\multirow{4}{*}{Study1}  & 1 style  &25.8 &71.1\\
 &10 style  &26.3 &71.8\\
 &25 style  &26.2 &71.7\\
 &50 style  &26.4 &71.5\\
 \midrule
\multirow{3}{*}{Study2} &No mask  &23.8 &61.6\\
 &Coarse map  &26.3 &71.8\\
 &Fine map  &27.6 &71.3\\

\bottomrule
\end{tabular}
\end{sc}
\end{scriptsize}
\end{center}

}
\hfill
\parbox{.5\linewidth}{
\caption{Mean AP scores on the KITTI benchmark. Easy (E), Moderate (M), and Hard (H).} \label{tbl::kitti}
\centering

\begin{center}
\begin{tiny}
\begin{sc}
\begin{tabular}{l|ccc|ccc|}
\toprule
\multicolumn{7}{c}{\texttt{GTA}$\rightarrow$\texttt{KITTI}}\\
\toprule
 &  \multicolumn{3}{c}{car} & \multicolumn{3}{c}{pedestrian}\\
%  \midrule
  &  E & M & H &   E & M & H\\
\midrule
\texttt{Synth}  & 70.4 &61.5 &51.0 & 45.5 &37.8 &34.2\\
\texttt{DR}  & 75.3 & 70.3 & 59.8 & 45.3& 39.3& 35.2 \\
\texttt{Cyclegan}   & 79.3&70.0&58.8 & 44.0& 38.3&34.9\\
\texttt{UNIT} & 76.6&68.2&56.9 & 48.6& 41.6 &37.5 \\
proposed   & \textbf{81.1}& \textbf{74.2} &\textbf{63.1} &\textbf{52.3} &\textbf{44.9}& \textbf{40.2} \\

\midrule
\texttt{Real}  & 90.6& 84.6& 75.3 & 67.4& 54.7&49.2 \\
\bottomrule
\end{tabular}
\end{sc}
\end{tiny}
\end{center}
}
\end{table}

\subsection{Synthetic-to-real Object Detection}

For object detection, we use the KITTI benchmark \cite{geiger2012we}, which consists of 7,481 labeled images captured by driving around the city of Karlsruhe. We split these images for 5-fold cross validation. The split is on video-sequence level. There is no overlapping video sequences between training and validation sets. We use the 2D bounding box annotations for car and pedestrian detection. For synthetic dataset, we use a set of images rendered from Grand Theft Auto V, tailored for the detection task. The setting is termed as \texttt{GTA}$\rightarrow$\texttt{KITTI}. For this task, we stylize the synthetic images without using any segmentation masks (e.g.\ $D_0^S$). We then train a detection network using the stylized synthetic images. 

We train a variant of the single shot detector \cite{liu2016ssd} for 100 epochs. We report the performance achieved in the last epoch averaged over 5 validation buckets. The performance is measured on car and pedestrian detection tasks under three different levels of difficulties, easy, moderate, and hard, as defined by the KITTI benchmark. 

Table~\ref{tbl::kitti} shows the results. We find that the proposed method improves the car detection mean AP score from 61.5 to 74.2 and pedestrian detection mAP score from 37.8 to 44.9 in the moderate setting, respectively. The improvement is clearly larger than those achieved by the GAN-based image translation methods.

\begin{figure}[t]
    \centering
    \begin{subfigure}[b]{0.66\textwidth}
        \centering
        \includegraphics[width=\linewidth]{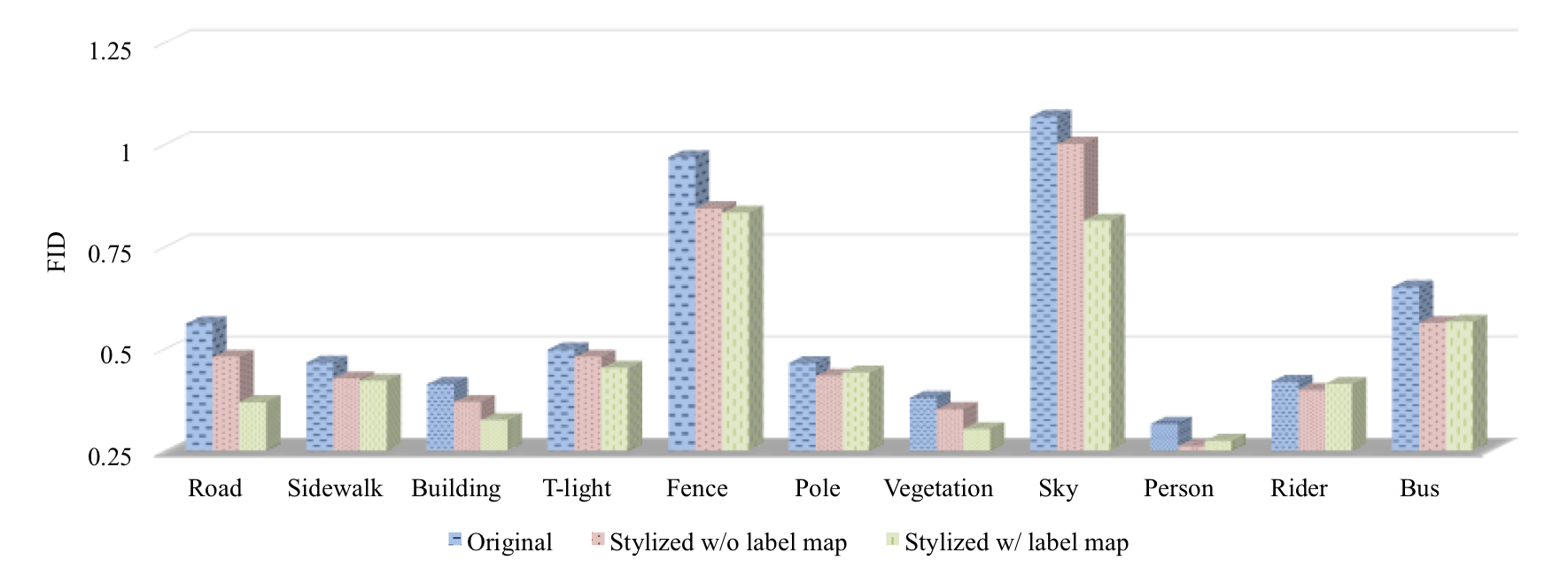}
       % \caption{FIDs for different classes}
        \label{fig::dist_perclass}
    \end{subfigure}%
    ~ 
    \begin{subfigure}[b]{0.33\textwidth}
        \centering
        \includegraphics[width=\linewidth]{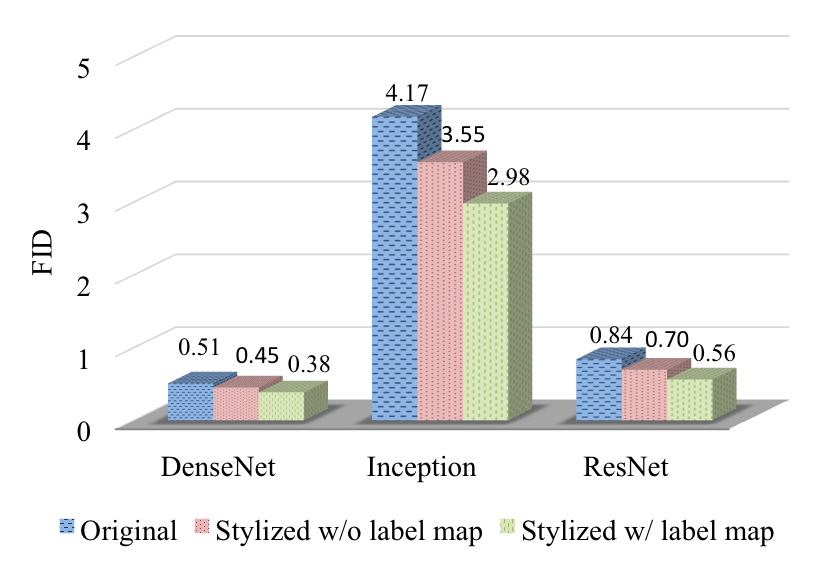}
        %\caption{Weighted average FIDs across all classes}
        \label{fig::dist_overall}
    \end{subfigure}
    \vspace{-7mm}
    \caption{Frechet Inception distances (\texttt{FID}s) between features of Cityscapes images and different variants of SYNTHIA images. (a) \texttt{FID}s of different classes on DenseNet161 layer \texttt{transition\_2}. (b) Weighted average \texttt{FID}s on feature extractors. It can be seen that domain stylization, especially with the aid of segmentation maps, helps reduce the distances.}
    \label{fig::dist_fid}
    %\vspace{-3mm}
\end{figure}

\section{Discussions}\label{sec::disc}

We conduct a study to explain effectiveness of domain stylization. We first note that as training with synthetic data but testing on real data, one faces the covariate shift problem~\cite{shimodaira2000improving}---the training data and test data follow different distributions but the conditional distributions of outputs given inputs are the same. As stylizing synthetic images with real images, we transform the distribution of synthetic data to better align with the distribution of real data. To quantify how well the two distributions align, we use the Frechet Inception distance (\texttt{FID})~\cite{heusel2017gans}. 

Suppose we have two datasets $D^A=\{(\mathbf{x}_A,\mathbf{m}_A)\}$, $D^B=\{(\mathbf{x}_B,\mathbf{m}_B)\}$. We first extract deep features, $\mathbf{f}^A$'s and $\mathbf{f}^B$'s, from $\mathbf{x}^A$'s and $\mathbf{x}^B$'s, respectively, using a feature extractor network derived from a pretrained network on the ImageNet classification task (the inception network). Next, we gather the features in $\mathbf{f}^A$'s whose semantic labels are $l$. This gives us a number of $C$-dimensional vectors, where $C$ is the feature dimension. We then fit a multi-variate Gaussian to these features vectors by computing the mean $\mathbf{\mu}_L^A$ and the co-variance matrix $\Sigma_L^A$ of these vectors. Similar, we have $\mathbf{\mu}_L^B$ and $\Sigma_L^B$. The \texttt{FID} between these extracted features is 
\begin{equation}
    \|\mathbf{\mu}_L^A - \mathbf{\mu}_L^B\|^2 + \text{Tr} \Big(\Sigma_L^A + \Sigma_L^B - 2 \sqrt{\Sigma_L^A \Sigma_L^B}\Big)
\end{equation}

In our experiment, we compute the \texttt{FID} between the Cityscapes dataset and three variants of the SYNTHIA dataset: the original images, domain stylized images without using label maps, and domain stylized images using predicted label maps. We use several different pretrained networks including DenseNet \cite{huang2017densely}, InceptionNet \cite{szegedy2016rethinking}, and ResNet \cite{he2015deep}. Figure~\ref{fig::dist_fid}(a) shows the FIDs for sample classes on DenseNet161 layer \texttt{transition\_2}. Figure~\ref{fig::dist_fid}(b) shows the average FIDs across all labels from different feature extractors, namely DenseNet161, InceptionV3, and ResNet101, weighted by class occurrence frequencies. It can be seen that when domain stylization is used, the distances between the synthetic images and real images are dramatically reduced. Moreover, the distances can be further reduced when the label maps are available during stylization.

\section{Conclusion}

We showed that by using an existing image style transfer algorithm, we can generate synthetic datasets to train deep neural networks for several semantic segmentation and visual object detection tasks, and that these networks yield better final performance than state-of-the-art image translation algorithms. Furthermore, our Frechet Inception distance based analysis justifies each step of our proposed approach by showing improved quantitative metrics in both stages of our algorithm. We expect the experimental results of this paper to serve as a baseline for future synthetic-to-real image translation algorithms, and we expect the results to improve as the image stylization algorithms develop.

%We showed that by using a simple, off-the-shelf image style transfer algorithm, we can create synthetic datasets to train deep neural networks for several semantic segmentation and visual object detection tasks that yield better final performance metrics. 

%Our proposed evaluation method can furthermore also serve as a useful evaluation metric for future work on image domain stylization algorithms. 

%trained with the enhanced synthetic dataset computed by adversarial training.

%\clearpage
%\clearpage
{\small
\bibliographystyle{ieee}
\bibliography{domain_stylization}
}

%\clearpage
%\appendix
%\input{appen}
\end{document}